\documentclass[letterpaper]{article} 
\usepackage{aaai2026}  
\usepackage{times}  
\usepackage{helvet}  
\usepackage{courier}  
\usepackage[hyphens]{url}  
\usepackage{graphicx} 
\usepackage{natbib}  
\usepackage{caption} 
\usepackage{algorithm}
\usepackage{algorithmic}
\usepackage{xspace}
\usepackage{bm}
\usepackage{multicol}
\usepackage{multirow}
\usepackage{makecell}
\usepackage{booktabs}
\usepackage{amsfonts}
\usepackage{amsmath}

\newcommand{\etal}{\emph{et\ al.}\xspace} 
\newcommand{\ie}{\emph{i.e.}\xspace}

\usepackage{newfloat}
\usepackage{listings}
\DeclareCaptionStyle{ruled}{labelfont=normalfont,labelsep=colon,strut=off} 
\floatstyle{ruled}
\newfloat{listing}{tb}{lst}{}
\floatname{listing}{Listing}
\title{Difficulty Controlled Diffusion Model for Synthesizing Effective Training Data}
\author {
    Zerun Wang\textsuperscript{\rm 1,\rm 2}\thanks{Work done during the internship at CyberAgent.},
    Jiafeng Mao\textsuperscript{\rm 1},
    Xueting Wang\textsuperscript{\rm 1}\thanks{Corresponding Author.},
    Toshihiko Yamasaki\textsuperscript{\rm 2}
}
\affiliations {
    \textsuperscript{\rm 1}CyberAgent \\
    \textsuperscript{\rm 2}The University of Tokyo \\
    \{ze\_wang, yamasaki\}@cvm.t.u-tokyo.ac.jp, \{jiafeng\_mao, wang\_xueting\}@cyberagent.co.jp
}

\begin{document}

\maketitle

\begin{abstract}
Generative models have become a powerful tool for synthesizing training data in computer vision tasks. Current approaches solely focus on aligning generated images with the target dataset distribution. As a result, they capture only the common features in the real dataset and mostly generate ``easy samples'', which are already well learned by models trained on real data. In contrast, those rare ``hard samples'', with atypical features but crucial for enhancing performance, cannot be effectively generated. Consequently, these approaches must synthesize large volumes of data to yield appreciable performance gains, yet the improvement remains limited. To overcome this limitation, we present a novel method that can learn to control the learning difficulty of samples during generation while also achieving domain alignment. Thus, it can efficiently generate valuable ``hard samples'' that yield significant performance improvements for target tasks. This is achieved by incorporating learning difficulty as an additional conditioning signal in generative models, together with a designed encoder structure and training–generation strategy. Experimental results across multiple datasets show that our method can achieve \textbf{higher performance} with \textbf{lower generation cost}. Specifically, we obtain the best performance with only 10\% additional synthetic data, saving 63.4 GPU hours of generation time compared to the previous SOTA on ImageNet. Moreover, our method provides insightful visualizations of category-specific hard factors, serving as a tool for analyzing datasets.
\end{abstract}

\begin{links}
    \link{Code}{https://github.com/komejisatori/Difficulty-Aware-Synthesis}
\end{links}

\section{Introduction}

Manually collecting and annotating a large number of images for training visual task models is time-consuming and labor-intensive. Recently, the rapid advancement of image generation models~\cite{dhariwal2021diffusion, ho2020denoising} offers a promising way to synthesize new training data automatically.

\begin{figure}[h!]
  \centering
  \includegraphics[width=1.0\linewidth]{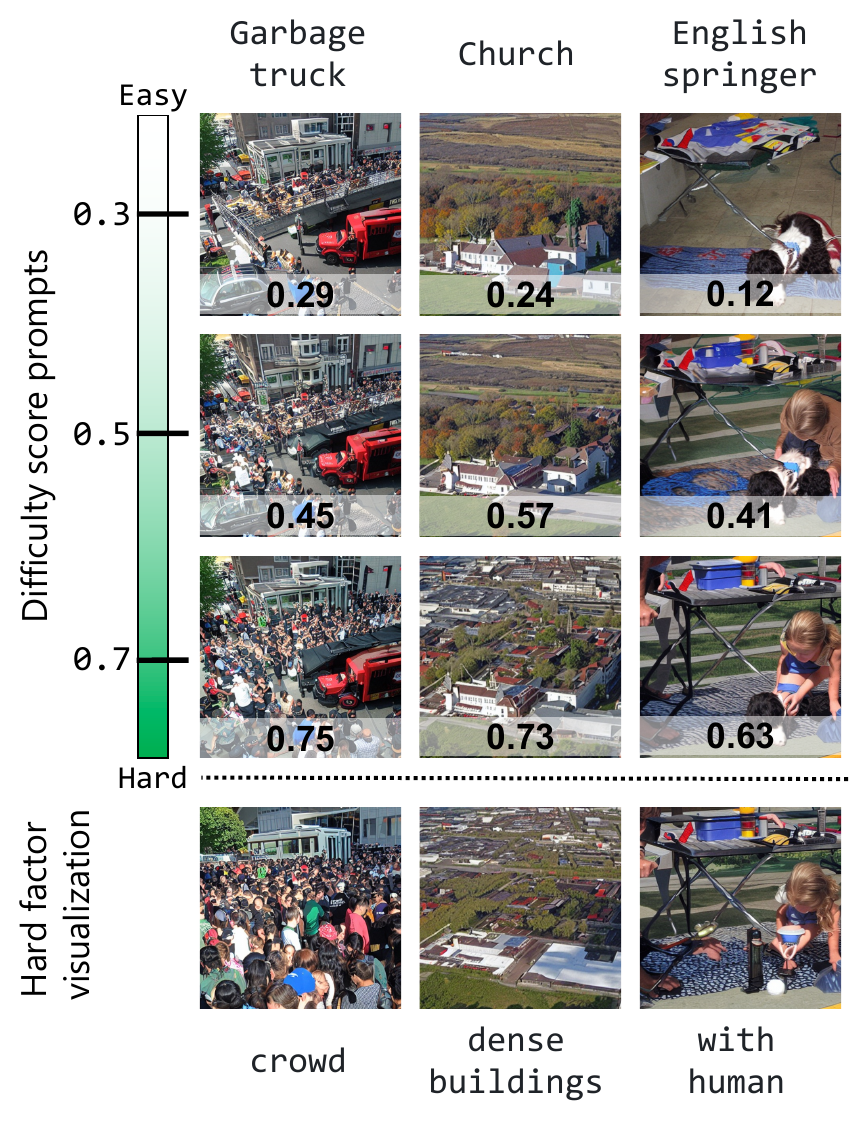}
  \caption{Our method generates images with controllable learning difficulty that align with specified difficulty score prompts. The score on each image is computed by a pretrained classifier. Additionally, our method reveals and visualizes the factors that contribute to sample difficulty.}
  \label{fig:fig1}
\end{figure}

Training data synthesis methods~\cite{sariyildiz2023fake,vendrow2023dataset,zhou2023training} have successfully enhanced model performance by augmenting the original datasets with images synthesized by generation models. A general pipeline is to use text-to-image generation models with text prompts related to target class names to generate training data.

Images generated by off-the-shelf diffusion models, however, often suffer from a distribution mismatch with the target dataset, reducing their effectiveness for training. Recent work~\cite{yuan2024realfake} addressed this issue by fine-tuning generation models on target real datasets to align distributions. However, this approach introduces new limitations: (1) The fine-tuned model tends to generate easy samples that reflect only the dominant features of the target dataset. These synthetic samples have low learning difficulty, as similar real images are already abundant in the dataset. Consequently, they contribute less to improving the target task. (2) The generation of hard samples is extremely limited. However, these samples, which have higher learning difficulty, are often more effective for enhancing target task performance because they contain minority or atypical features of the target dataset. Thus, a dilemma arises in current methods: without fine-tuning, generated images suffer from distribution mismatch, while aligning domains through fine-tuning leads to primarily generating samples with low learning difficulty.

We address this dilemma by introducing learning difficulty as an additional conditioning signal during fine-tuning and generation. The model can thus capture the features of samples with different learning difficulties while still maintaining domain alignment. We fine-tune the pretrained text-to-image model using both difficulty score prompts and text prompts as conditioning inputs. Then, the model can generate new samples at specified difficulty levels by adjusting the input difficulty score prompt. Our approach can disentangle learning difficulty from the domain alignment process, enabling the generation of samples with varying difficulties while maintaining domain consistency.

Extensive experiments across multiple image classification datasets demonstrate the effectiveness of our proposed method in generating samples with controllable learning difficulty, thereby improving the performance of the target task more efficiently. Furthermore, our method enables the visual analysis of class-wise hard factors, providing insights into what makes certain class samples difficult in the target dataset. Our contribution can be summarized as follows,

\begin{itemize}
    \item We show that domain alignment alone causes models to generate mostly easy samples, which provide only limited performance gains.  
    \item We propose a difficulty-controlled generation framework that can be used to synthesize samples with targeted learning difficulty, thus improving performance on target tasks more efficiently.
    \item We validate our method across multiple image classification tasks, demonstrating its effectiveness in both providing valuable training samples and capturing factors that affect sample difficulty for visualization.
\end{itemize}

\section{Related Work}
\label{sec2}
\subsection{Conditioned Image Generation}
Diffusion models are one of the mainstream tools for generating images~\cite{dhariwal2021diffusion, ho2020denoising, ho2021classifier, nichol2021improved, song2020denoising, liu2022pseudo}. Starting from Gaussian noise, these models iteratively predict the noise to be removed at each step, gradually denoising samples to obtain high-fidelity outputs. Many contemporary methods~\cite{kim2022diffusionclip, ramesh2021zero, ding2021cogview, gafni2022make} leverage text prompts encoded by CLIP~\cite{radford2021learning} to guide the denoising process. Notably, Latent Diffusion~\cite{rombach2022high} conducts denoising in a latent space before decoding the denoised latents into pixel space. Various approaches have been explored to enhance control over the generation process through different conditioning signals, such as image-based guidance~\cite{mou2023t2i, kawar2023imagic, ruiz2023dreambooth, brooks2023instructpix2pix, avrahami2022blended, wang2022pretraining}, compositional conditioning~\cite{liu2022compositional, park2021benchmark, huang2023composer}, and layout guidance~\cite{mao2023guided, zhang2023adding, li2023gligen, voynov2022sketch}. In contrast to these works, we propose a novel form of generation guidance using a score reflecting the learning difficulty of training samples.

\subsection{Training Data Synthesis}
Synthetic images have proven effective for serving as new training data in deep learning vision tasks, thereby saving the cost of collecting and labeling real-world data. Early works~\cite{zhang2021datasetgan, besnier2020dataset} utilize generative adversarial network (GAN)-based models. Recent works apply more powerful diffusion models for data synthesis. Several approaches directly utilize off-the-shelf pre-trained diffusion models: Sariyildiz \etal~\cite{sariyildiz2023fake} applied text prompt engineering strategies to improve the diversity of generated results. Huang \etal~\cite{huang2024active} augment misclassified real data by using them as image guidance for the diffusion model. Meanwhile, some works~\cite{vendrow2023dataset,zhou2023training} leverage textual inversion techniques to encode class-specific characteristics from real data into new tokens. Recent methods point out the importance of aligning the distribution between synthetic and real data. This is achieved by fine-tuning the diffusion models using real data. Azizi \etal~\cite{azizi2023synthetic} finetuned the Imagen~\cite{saharia2022photorealistic} model for data generation. Real-Fake~\cite{yuan2024realfake} theoretically shows that fine-tuning achieves domain alignment. They apply the more effective Low-Rank Adaptation (LoRA)~\cite{hu2022lora} approach to fine-tune the Stable Diffusion~\cite{rombach2022high} model. These fine-tuning methods, however, tend to reproduce dominant features of the target dataset and consequently generate mostly easy samples. In contrast, our method can generate samples with appropriate difficulty in the target domain, thus further improving the performance.

\section{Preliminary}
In this section, we first introduce the setting of training data synthesis. Then, we introduce our investigation of the dilemma of previous training data synthesis methods.

\subsection{Task Setting}

Following previous methods~\cite{azizi2023synthetic,yuan2024realfake}, we evaluate our method on basic image classification tasks. Training data synthesis leverages generative models, typically text-to-image diffusion models, to augment a target dataset $\mathcal{D}_t = \{\bm{x_i}, y_i\}^{n_t}_{i=1}$ with synthetic images and labels $\mathcal{D}_s = \{\bm{x_i^\prime}, y^\prime_i\}^{n_s}_{i=1}$, where $\bm{x_i}$ and $\bm{x_i^\prime}$ denote the real and synthetic images. $y_i$, $y^\prime_i$ denote the labels for the classification task. The combined dataset $\mathcal{D} = \mathcal{D}_t \cup \mathcal{D}_s$ can be used to train a target classification model, which typically outperforms models trained solely on $\mathcal{D}_t$.

\subsection{Investigation}
\label{sec3_2}
Previous methods aligned the domain of $\mathcal{D}_s$ with $\mathcal{D}_t$ by fine-tuning the generative model. However, we investigated their generated data from the learning difficulty perspective and found that they are dominated by ``easy samples''.

\paragraph{Difficulty Score.} We define a difficulty score $s$ to represent the learning difficulty of a sample. Given $c\in(0, 1)$ as the predicted probability of the ground-truth class produced by a classifier after the softmax activation, the score is computed as
\begin{equation}
    s = 1-c.
\end{equation}
Thus, a higher difficulty score leads to a sample with higher learning difficulty for the model and vice versa. 

\begin{table}[t]
\centering
\begin{center}
\resizebox{\linewidth}{!}{
\begin{tabular}{c|c|ccc}
\toprule
\multirow{2}{*}{Data} & \multirow{2}{*}{Real only} &  \multicolumn{3}{c}{+ Synthetic data} \\
\cmidrule(lr){3-5}
   ~ &     & Easy & Medium & Extremely hard\\
\midrule
Acc.  &       $95.0$ & $95.2\ (+0.2)$ & $\bm{95.8\ (+0.8)}$ & $94.6\ (-0.4)$ \\

\bottomrule
\end{tabular}
}
\end{center}

\caption{Classification accuracy when real data is augmented with synthetic data from different difficulty score ranges.}

\label{tab:tab1}
\end{table}

\begin{figure}
  \centering
  \includegraphics[width=0.75\linewidth]{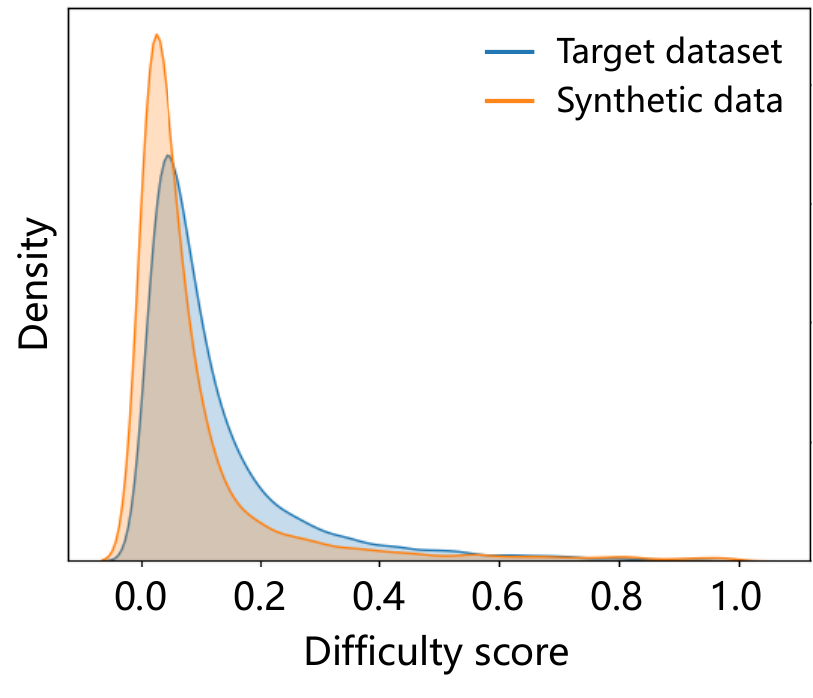}
  \caption{KDE distribution curve of difficulty scores. Simply fine-tuning on the whole target dataset biases the model toward generating easy images.}
  \label{fig:fig2}
\end{figure}

\paragraph{Dilemma of Current Methods.} We then analyzed the difficulty score distribution and its impact on target-task performance using the recent Real-Fake method, and found that the generated samples are dominated by easy examples.
We used Imagenette~\cite{Howard_Imagenette_2019} as the target classification dataset and applied a ResNet-50~\cite{he2016deep} model pretrained on the training split to compute the difficulty score. Based on our experiments, we found that:

\noindent\textbf{(1) Current methods mainly generate easy samples.} We first compared the difficulty score distribution of $\mathcal{D}_t$ and $\mathcal{D}_s$ generated by Real-Fake. We generated the same number of images as the real training dataset and assessed the difficulty score. The Kernel Density Estimation (KDE) distributions in Fig.~\ref{fig:fig2} show that the synthetic data distribution is more extreme and dominated by easy samples.

\noindent\textbf{(2) Samples with appropriate difficulty are more effective.} We further analyzed the relationship between the difficulty score and the improvement effect on the target task. We divided the difficulty score $s\in(0.0, 1.0)$ into three levels: Easy $(0, 0.33)$, Medium  $(0.33, 0.66)$, and Extremely Hard$(0.66, 1)$. Then we generated and selected the same amount of data (25\% of the real training set) for each difficulty level. The results in Table~\ref{tab:tab1} show that synthetic data with a medium-level difficulty score yields the most significant training improvements. However, Real-Fake is \textbf{highly inefficient} at generating such samples as the medium-difficulty examples constitute only about 1\% of all generated images (Fig.~\ref{fig:fig2}). As a result, a large amount of additional data needs to be generated to filter out a sufficient number of medium-difficulty samples. Meanwhile, using only extremely hard samples degrades training performance, highlighting the importance of controllable sample difficulty. 

\section{Difficulty Controlled Dataset Synthesis}
\begin{figure*}[t]
  \centering
  \includegraphics[width=0.99\linewidth]{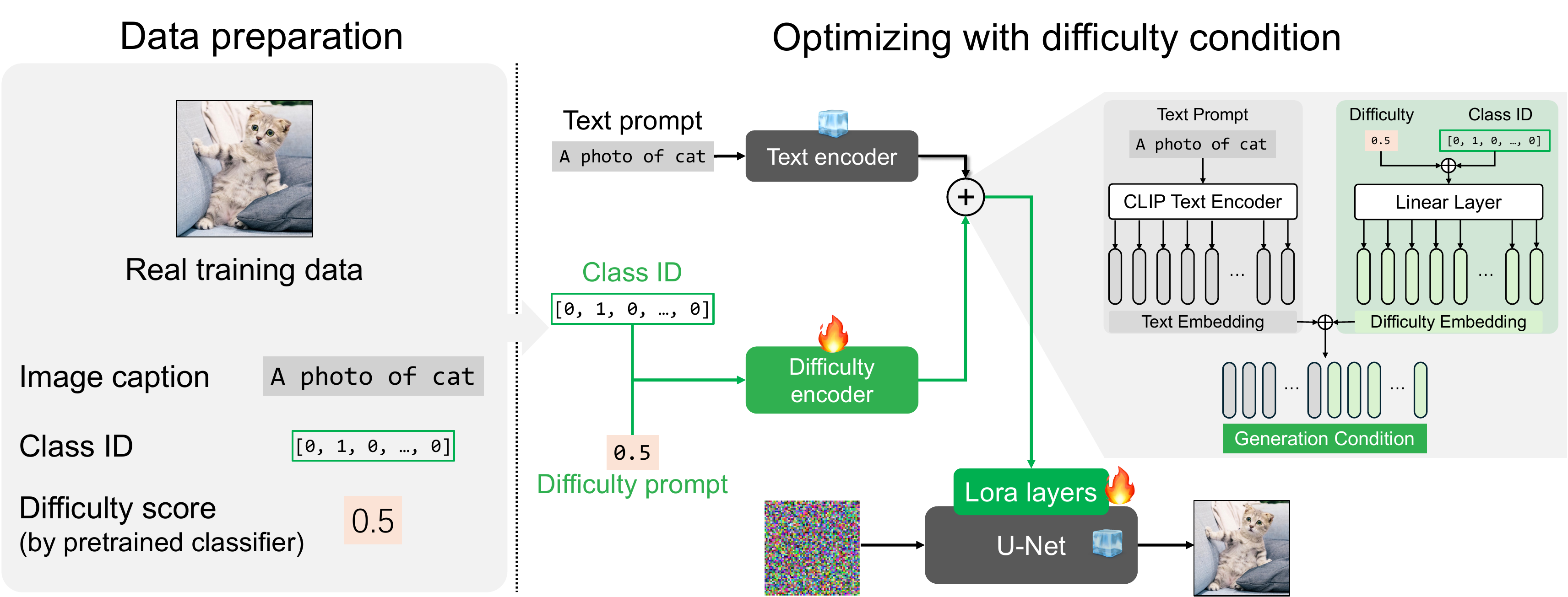}
  \caption{Overview of our method. \textbf{Left}: Real training images are annotated with a text caption and a difficulty score assessed by a pretrained classifier. \textbf{Right}: A difficulty encoder is integrated into the text-to-image diffusion model. The model is fine-tuned to incorporate the difficulty score as an additional condition.}
  \label{fig:overview}
\end{figure*}

Previous methods suffer from the above dilemma because their strategies, while achieving domain alignment, captured only the dominant features of the target dataset. Motivated by this observation, we propose a method that disentangles (1) domain alignment and (2) difficulty-aware feature modeling. We achieve this by introducing learning difficulty as an explicit conditioning to control the image generation process. Fig.~\ref{fig:overview} illustrates the structure of our model and the fine-tuning pipeline. For the model structure, we incorporate a difficulty encoder into a standard text-to-image diffusion model. During fine-tuning, the model is trained on data annotated with difficulty scores and text prompts. With our method, the learning difficulty of generated images can be controlled by adjusting the input difficulty score prompt.

\subsection{Model Structure}
We adopt the text-to-image Stable Diffusion~\cite{rombach2022high} model for image generation. The model consists of a CLIP text encoder and a denoising U-Net. Additionally, we introduce a difficulty encoder to condition the model on the input difficulty score.

\paragraph{Difficulty Encoder.} To establish the mapping relationship between difficulty scores and the characteristics of samples, we construct a difficulty encoder, $\mathcal{E}_d$, which is a multilayer perceptron (MLP) model that projects the difficulty score of the $i$th sample into a latent embedding $\bm{h_i}$ for controlling the generation. 

Although samples may share the same difficulty score, their visual characteristics can differ substantially across categories. Thus, the difficulty encoder must produce distinct embeddings for different categories. Therefore, our difficulty encoder takes as input the concatenation of the category labels and the difficulty score, \ie, $\bm{h_i} = \mathcal{E}_d( [y_i] \oplus [s_i] )$, where $\oplus$ indicates a concatenation operation. 

$\bm{h_i}$ is concatenated with the CLIP text-prompt embedding produced by the pretrained CLIP encoder $\mathcal{E}_{\text{text}}$ in the Stable Diffusion model, as illustrated in Fig.~\ref{fig:overview}. We use the concatenated embeddings, $\bm{\tau}_i = \mathcal{E}_\text{text}\left(p_i\right) \oplus \bm{h_i}$, to guide the generation process, enabling control based on both the text description $p$ and the difficulty score.

\subsection{Difficulty Controlled Fine-Tuning}
\label{sec:diffusion_finetune}

\paragraph{Data Preparation.} 
To fine-tune the text-to-image model on the target dataset $\mathcal{D}_t$ using our method, we generate difficulty scores and text prompts for each training sample.

For difficulty score, we use a classifier pretrained on $\mathcal{D}_t$ to compute each sample's score $s$, as defined in the \textit{Investigation section}. Note that we do not depend on any specific classifier architecture for scoring. For text prompts, each image is paired with a text caption $p$ in the form of ``\texttt{a photo of [CLS]}'', where \texttt{[CLS]} denotes the category name in the target dataset. In contrast to previous works using complex prompts generated by pretrained captioning models, we adopt a simple template during training, enabling difficulty control to be handled by the difficulty encoder rather than by prompt complexity.

Finally, the original target dataset $\mathcal{D}_t$ is extended to $\mathcal{D}_t^\prime = \{\bm{x}_i, y_i, s_i, p_i\}^{n_t}_{i=1}$, where $s_i$ and $p_i$ denote the assigned difficulty score and text prompt for the image.

\paragraph{Optimization with Difficulty Condition.} The difficulty encoder is trained from scratch. And we employ Low-Rank Adaptation (LoRA)~\cite{hu2022lora} to efficiently fine-tune the diffusion model on our prepared dataset. The model is trained to predict the added noise $\epsilon$ given a noised latent $\bm{z}_t$ at timestep $t \in \{1, \dots, T\}$, using the conditioning input. The denoising loss is formulated as:
\begin{equation}
    \mathcal{L} = \mathbb{E}_{\mathcal{E}(\bm{x}), \bm{\tau}, \epsilon \sim \mathcal{N}(0,1),t} \left[\parallel \epsilon - \epsilon_{(\theta,\delta)}(\bm{z}_t, t, \bm{\tau})\parallel_2^2 \right],
\label{eq:1}
\end{equation}
where $\delta$ denotes the parameters of LoRA layers. 

During the training process, LoRA enables efficient fine-tuning of the diffusion model to denoise perturbed images, ensuring the generated outputs align with the target dataset distribution. Simultaneously, the difficulty encoder learns to project difficulty scores into a latent space to control the generation process. The distinct roles of these two learnable components effectively disentangle the domain alignment and difficulty control.

\subsection{Difficulty Controlled Data Synthesis}
After optimizing on the target dataset, we employ the model for training data synthesis. For each class in the target classification task, we use class-specific text prompts and sample difficulty scores from a predefined distribution.

For text prompts, unlike during training, where simple templates are used, we adopt more diverse text descriptions, following Real-Fake, to enhance generation diversity. For the difficulty score distribution, we sample the difficulty score $s$ from a Gaussian distribution $s \sim \mathcal{N}(\mu, \sigma)$, where $\mu$ is centered around a mid-level difficulty, and the standard deviation $\sigma$ controls the degree of diversity. We experimentally found the suitable $\mu, \sigma$ during generation.

The generated images are then used to augment the target dataset and enhance classification performance. We experimentally demonstrate that our method effectively controls difficulty while remaining compatible with diverse text prompts.

\begin{figure*}[t]
  \centering
  \includegraphics[width=1\linewidth]{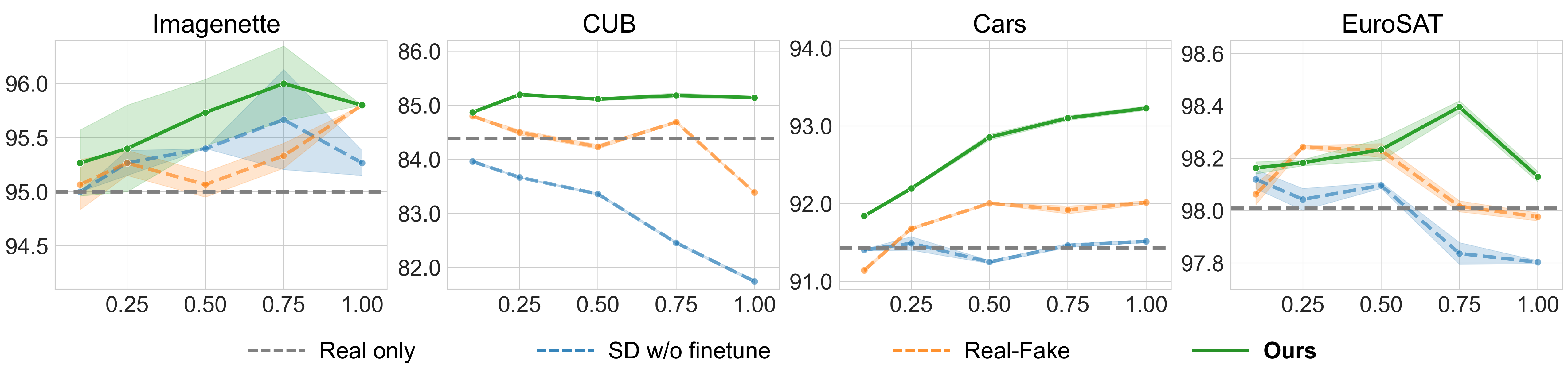}
  \caption{Top-1 classification accuracy (\%) on various tasks with ResNet-50 model. The x-axis denotes the ratio of additional synthetic images. All results are averaged over three runs, and shaded regions represent the standard deviation. The detailed numerical results are provided in the appendix.}
  \label{fig:results}
\end{figure*}

\begin{table*}[t]
\begin{center}

\begin{tabular}{lccccccc}

\toprule
\multirow{2}{*}{Method}  &  \multicolumn{7}{c}{Synthetic data ratio} \\
\cmidrule(lr){2-8}
     & Real only & 5\% &  10\%   & 25\%     &  50\% & 75\% & 100\%  \\
\midrule
Generation time (GPU hours) & 0 & 7.9 & 15.9 & 39.6 & 79.3 & 118.9 & 158.5 \\
\midrule
SD w/o finetune  & \multirow{3}{*}{$78.21$} & $77.96$  & $77.90$      &  $77.80$ &  $77.91$  &  $77.73$  &  $77.64$  \\
Real-Fake~\cite{yuan2024realfake}  &   & $78.32$  & $78.61$    &  $78.68$ &  $78.73$  &  $78.70$  &  $78.62$   \\
\textbf{Ours}  & & $\bm{78.47}$ & $\bm{78.74}$ & $\bm{78.76}$ &  $\bm{78.73}$  &  $\bm{78.71}$ & $\bm{78.63}$    \\
\bottomrule
\end{tabular}
    
\caption{Top-1 classification accuracy (\%) on ImageNet with ResNet-50. GPU hours are measured on a single NVIDIA A100 GPU with a generation batch size of 128. The difficulty encoder introduces an additional latency of only 8\%. Therefore, all reported timings are based on the original Stable Diffusion model.}
\label{tab:tab2}
\end{center}
\end{table*}

\begin{table}[t]
\centering
\begin{center}
\begin{tabular}{l|ccccc}
\toprule
& \multicolumn{4}{c}{Standard deviation $\sigma=0.1$} \\
\midrule
Mean value ($\mu)$ & $0.3$ & $0.5$ & $0.7$ & $0.9$  \\
\midrule
Acc.   & $95.8$ & \bm{$96.4$} & $ 96.0 $ & $95.2$   \\
\bottomrule
\toprule
& \multicolumn{4}{c}{Mean value $\mu=0.5$} \\
\midrule
Standard deviation ($\sigma)$ & 0.00 & 0.01 & 0.10 & 0.50 \\
\midrule
Acc. & $95.8$ & $95.8$ & $\bm{96.4}$ & $96.0$ \\

\bottomrule
\end{tabular}
\end{center}

\caption{Classification accuracy on Imagenette under different difficulty score distributions. A synthetic data ratio of 75\% is used.}

\label{tab:tab3}
\end{table}
\begin{table}[t]
\centering
\begin{center}
\begin{tabular}{l|ccc}
\toprule
Model & ViT-small & ResNet-50 & ResNet-101  \\
\midrule
Real only & $82.6$ & $95.0$ & $95.6$  \\
Real-Fake & $84.8$ & $95.4$ & $95.8$  \\
Ours & $\bm{86.0}$ & $\bm{96.4}$ & $\bm{96.8}$  \\
\bottomrule
\end{tabular}
\end{center}

\caption{Classification accuracy of different model structures on Imagenette. A synthetic data ratio of 75\% is used.}

\label{tab:tab4}
\end{table}

\subsection{Hard Factors Extraction} 
\label{sec4_4}
Our difficulty encoder generates latent embeddings corresponding to specific levels of learning difficulty. This allows our method to analyze difficulty-inducing visual factors inherent in the target dataset. To achieve this, we generate images conditioned solely on difficulty scores, without using any text prompts, thereby isolating the visual features associated with different levels of difficulty. We demonstrate this analysis in our experiments.

\section{Experiments}

\subsection{Implementation Details}
\label{sec:exp_detail}
\paragraph{Target Tasks and Models.} We use Imagenette~\cite{Howard_Imagenette_2019}, CUB~\cite{wah_2023_cvm3y-5hh21}, Cars~\cite{krause20133d}, and ImageNet~\cite{deng2009imagenet} as target taskss. For the target classifiers, we use ResNet-50, ResNet-101~\cite{he2016deep}, and ViT-Small~\cite{dosovitskiy2020image}. The training settings follow Real-Fake~\cite{yuan2024realfake} for a fair comparison. 

\paragraph{Diffusion Model.} We use the pretrained Stable Diffusion v1.5~\cite{rombach2022high} with an image resolution of $512\times512$, following the same configuration as Real-Fake. Our implementation is based on the Diffusers library~\cite{von-platen-etal-2022-diffusers} library, and the LoRA~\cite{hu2022lora} fine-tuning schedule follows its default configuration.

\paragraph{Training Data Synthesis.} We augment the real training data for each classification task by generating synthetic images at various ratios. Generation is performed using the default PLMS sampler with 30 steps and a guidance scale of $2.0$. Following Real-Fake, we use captions generated by the pretrained BLIP-2 model~\cite{li2023blip} for Imagenette and ImageNet, and simple prompts of the form ``\texttt{a photo of [CLS]}'' for the other datasets. Difficulty scores are sampled from a Gaussian distribution with mean $\mu = 0.5$ and standard deviation $\sigma = 0.1$. Other detailed settings are reported in the appendix.

\begin{figure}[t]
  \centering
  \includegraphics[width=1\linewidth]{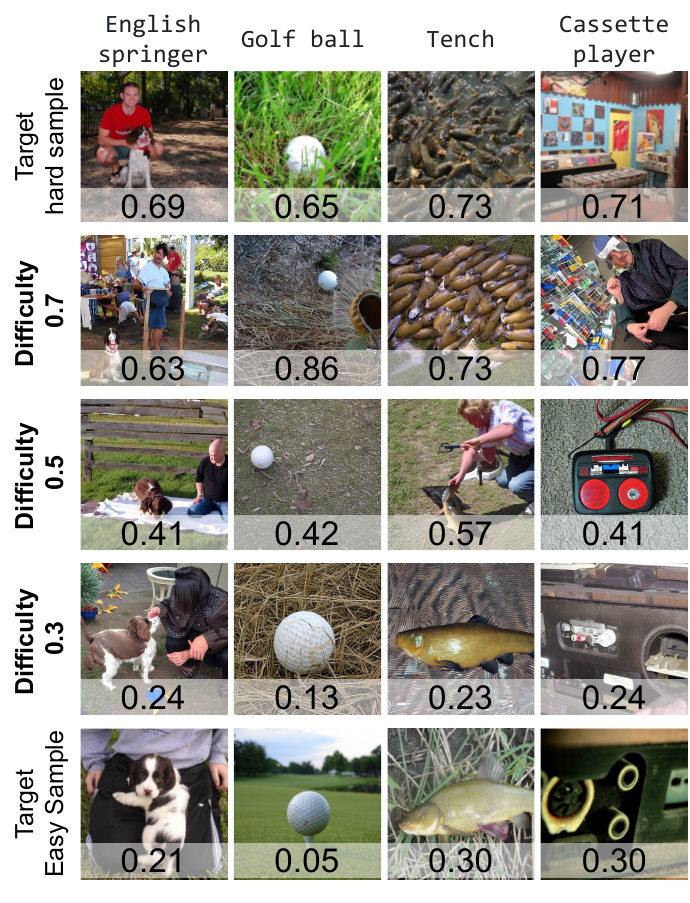}
  \caption{Visualization of synthetic images from four classes in Imagenette, with different difficulty score inputs shown on the left. Easy and hard samples from the target dataset are also shown for comparison. The numbers on the images are difficulty scores assessed by a pretrained ResNet-50 model.}
  \label{fig:quality_sample}
\end{figure}

\begin{figure}[t]
  \centering
  \includegraphics[width=1\linewidth]{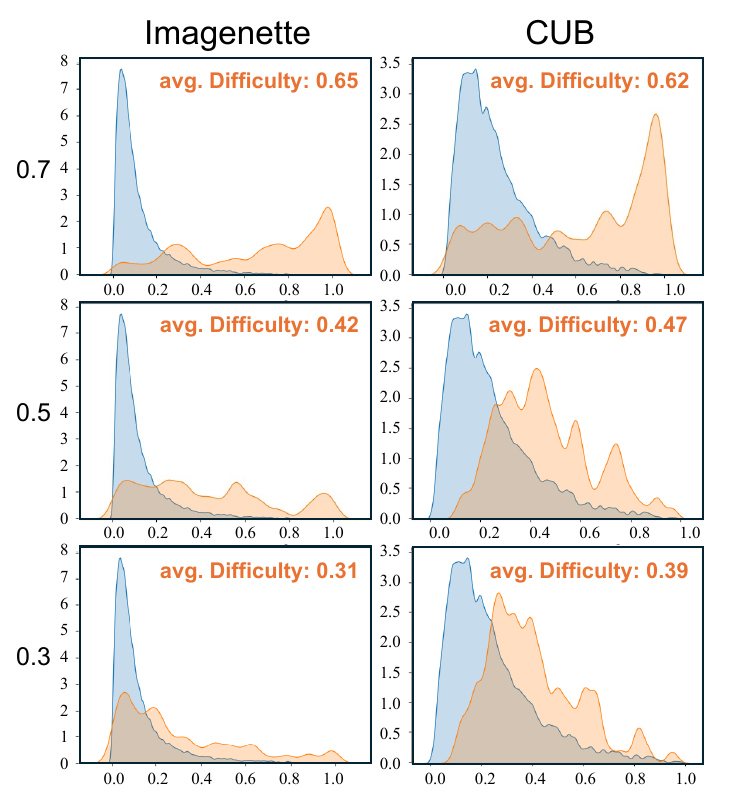}
  \caption{Difficulty score distributions of the real datasets (blue) and 100 randomly generated images (orange) using our method, shown for two datasets. X-axis: difficulty score; Y-axis: KDE density. Each row corresponds to a different difficulty score input, as indicated on the left. All difficulty scores are assessed by a pretrained ResNet-50 model.}
  \label{fig:score_dist}
\end{figure}

\begin{figure}[t!]
\begin{center}
\includegraphics[width=0.85\linewidth]{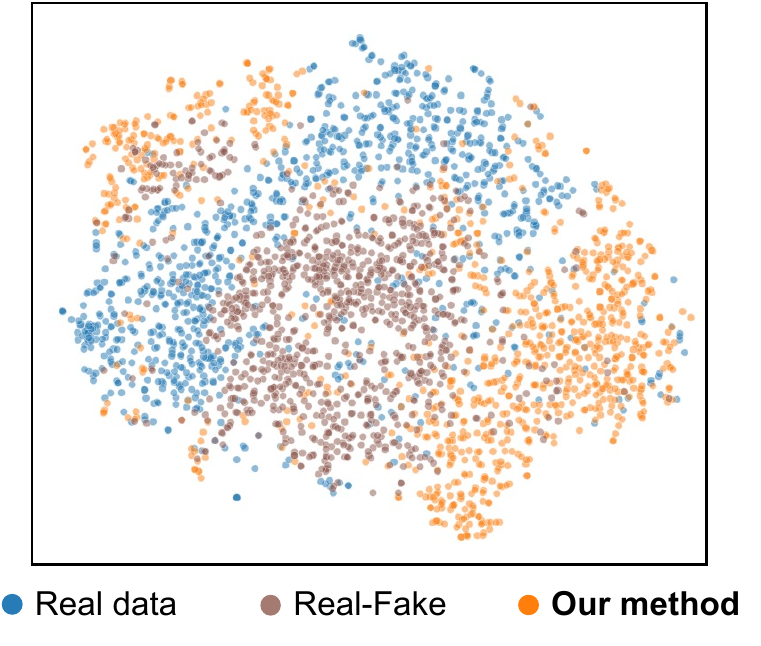}
\label{fig:hard_tsne}
\caption{T-SNE visualization of features extracted by a pretrained ResNet-50 for samples from the same class in Imagenette.}
\label{fig:tsne} 
\end{center}
\end{figure}

\begin{figure}[t]
  \centering
  \includegraphics[width=1\linewidth]{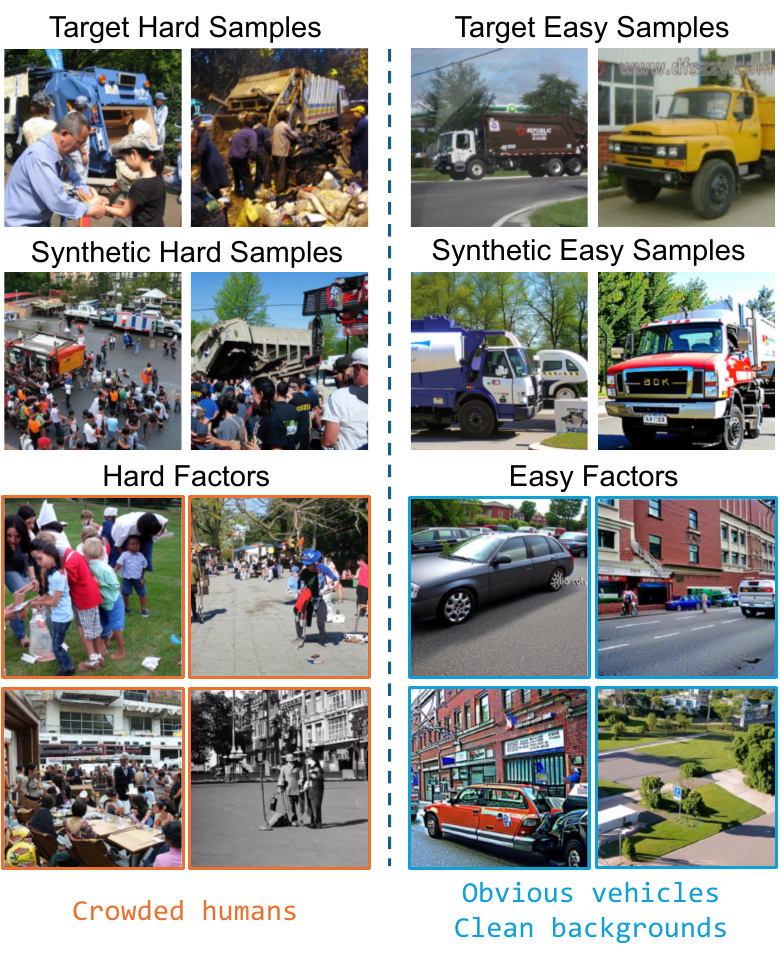}
  \caption{Hard factor visualization for the ``garbage truck'' category in Imagenette. Our method captures both hard and easy factors associated with the sample's learning difficulty.}
  \label{fig:hard_factor}
\end{figure}

\begin{figure}[t]
  \centering
  \includegraphics[width=1\linewidth]{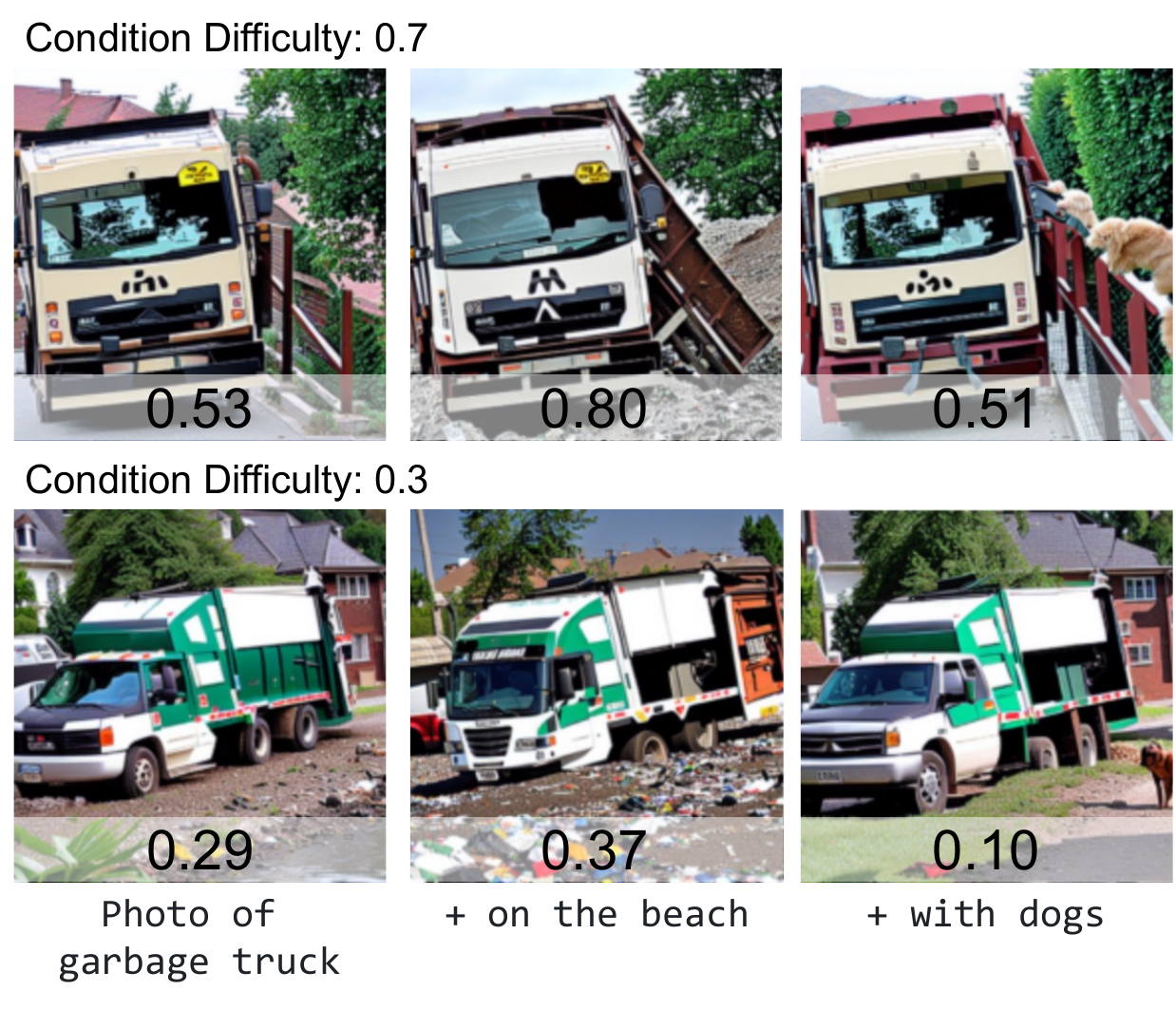}
  \caption{Our method supports diverse text prompts and can reflect both textual semantics and the target difficulty score.}
  \label{fig:compatibility}
\end{figure}

\subsection{Training Performance of Synthetic Data}
\label{sec:exp_classifier}

\paragraph{Classification Accuracy.} To evaluate the effectiveness of our method, we augment the real training split in each task with varying amounts of synthetic data generated by three methods: (1) our proposed approach, (2) Real-Fake, and (3) a Stable Diffusion model without fine-tuning. Fig.~\ref{fig:results} presents results on Imagenette, CUB, Cars, and EuroSAT. The results indicate that \textbf{our method achieves higher accuracy using fewer synthetic samples.} For example, on the Cars dataset, our method improves accuracy by over 1.2\% compared to Real-Fake. Table~\ref{tab:tab2} reports the results on ImageNet, showing that generating only an additional 10\% of data with our method is sufficient to surpass Real-Fake’s best result, while significantly reducing generation cost by approximately 63.4 GPU-hours of generation time. Moreover, compared to other methods, our method achieves more stable improvements over training with real data only.

\paragraph{Parameter Optimization.} 
We further analyze the hyperparameters used in our data synthesis strategy. Table~\ref{tab:tab3} shows the effect of the difficulty score distribution parameters ($\mu, \sigma$) on classification accuracy. The results show that our method enables effective tailoring of the synthetic data to each target dataset by simply tuning the difficulty score distribution. Moreover, our method exhibits robustness to parameter selection, consistently outperforming Real-Fake across a broad range of $\mu$ and $\sigma$ values.

\paragraph{Various Model Structures.} 
Table~\ref{tab:tab4} shows the effectiveness of our method on various model structures, including CNN and Transformer-based models. All synthetic data are generated using difficulty scores assessed by a ResNet-50 model pretrained on the target dataset. This demonstrates that our method does not rely on using the same model architecture for data preparation and final training.

\subsection{Controlling Efficacy of Difficulty}
\label{sec:exp_generate}

\paragraph{Visualization.} We randomly generate $100$ samples with each of the difficulty scores $0.3$, $0.5$, and $0.7$. Representative examples of generated images are shown in Fig.~\ref{fig:quality_sample}. The generated samples exhibit visual properties consistent with those of real data at corresponding difficulty levels.

\paragraph{Difficulty Score Distribution.} Fig.~\ref{fig:score_dist} shows the difficulty score distributions of the generated images. This demonstrates that our method effectively controls the learning difficulty of generated data, addressing a key limitation of existing methods that tend to produce mostly easy samples.

\paragraph{Feature Distribution.} We compare the feature distributions of images generated by our method and Real-Fake. As shown in Fig.~\ref{fig:tsne}, our method generates samples that more effectively delineate the boundaries of the feature distribution of real data, whereas Real-Fake's samples tend to cluster together. This indicates that our method effectively disentangles difficulty control from the domain alignment process.

\subsection{More Analysis}
 
\paragraph{Hard Factor Visualization.} Omitting the text prompt during generation allows our method to reveal the visual factors associated with different difficulty levels for the target dataset. As shown in Fig.~\ref{fig:hard_factor}, generated samples in the ``garbage truck'' category reveal that hard examples typically depict crowded scenes, while easy ones feature clearly visible vehicles and uncluttered backgrounds. Similar analyses for other categories are provided in the appendix.

\paragraph{Compatibility with Diverse Text Prompts.} Despite being fine-tuned on simple text prompts, our method generalizes well to more diverse inputs, as illustrated in Fig.~\ref{fig:compatibility}. The results show that our method successfully aligns with both difficulty scores and diverse text prompts. This enables the integration of our method with captions generated by powerful vision-language models~\cite{lei2023image}, enabling the synthesis of more diverse images.

\section{Conclusions}

In this paper, we demonstrate the critical role of appropriately difficult samples in model training. We identify a fundamental limitation in existing training data synthesis methods: domain alignment alone causes models to capture only common features and to generate mostly easy samples. To overcome this, we propose a generation method that conditions on learning difficulty to disentangle difficulty-aware feature modeling from domain alignment. We validate the effectiveness of our method through extensive experiments by evaluating classification performance and the difficulty distributions of the generated samples. In addition, our method serves as a useful tool for the visual analysis of hard factors within datasets.


\bibliography{aaai2026}

\end{document}